\documentclass[11pt]{article}

\usepackage[preprint]{acl}

\usepackage{times}
\usepackage{latexsym}

\usepackage[T1]{fontenc}

\usepackage[utf8]{inputenc}

\usepackage{microtype}

\usepackage{inconsolata}

\usepackage{graphicx}

\usepackage{hyperref}

\usepackage{fancybox}

\usepackage{tcolorbox}
\usepackage{subfig}
\usepackage{paralist}
\usepackage{enumitem}
\usepackage{xcolor}
\usepackage{cleveref}
\usepackage{soul}
\usepackage[normalem]{ulem}

\title{\textsc{AnimatedLLM}: Explaining LLMs with Interactive Visualizations}

\author{Zdeněk Kasner \and Ondřej Dušek \\
  Charles University \\
  Faculty of Mathematics and Physics \\
  Institute of Formal and Applied Linguistics \\
  \texttt{\{kasner,odusek\}@ufal.mff.cuni.cz}}

\begin{document}
\maketitle
\begin{abstract}
Large language models (LLMs) are becoming central to natural language processing education, yet materials showing their mechanics are sparse. We present \textsc{AnimatedLLM}, an interactive web application that provides step-by-step visualizations of a Transformer language model. \textsc{AnimatedLLM} runs entirely in the browser, using pre-computed traces of open LLMs applied on manually curated inputs. The application is available at \linebreak \url{https://animatedllm.github.io}, both as a teaching aid and for self-educational purposes.\footnote{The source code is available under the MIT license at \url{https://github.com/kasnerz/animated-llm}.}
\end{abstract}

%

\section{Introduction}

Natural language processing (NLP) researchers and teachers often need to explain the inner workings of large language models (LLMs) to non-experts. However, describing certain concepts -- for example, how the words are tokenized or how the model consumes its own output to generate coherent text -- is difficult without visual aids.

Existing visualization tools fall into two categories: static educational materials, which are useful for self-study but lack interactivity \citep{alammar2018illustrated,alammar2019illustrated,3blue1brown}, or tools for understanding the details of the Transformer architecture, which are directed towards technical audiences \citep{bycroftllm, cho2024transformer, brasoveanu2024visualizing}.

To bridge the gap between the existing tools, we present \textsc{AnimatedLLM}: a client-side web application designed for explaining the mechanics of LLMs to non-technical audiences. The goal of \textsc{AnimatedLLM} is to provide the right level of abstraction between the two extremes -- the complex matrix operations driving the Transformer architecture on the one hand, and the clean but opaque interface of AI assistants on the other.
Our learning objective is demystifying the LLM architecture and providing a high-level understanding of their inner workings: tokenization, embeddings, attention, autoregressive generation and next-word prediction training.

\begin{figure}[t]
  \centering
  \includegraphics[width=\columnwidth]{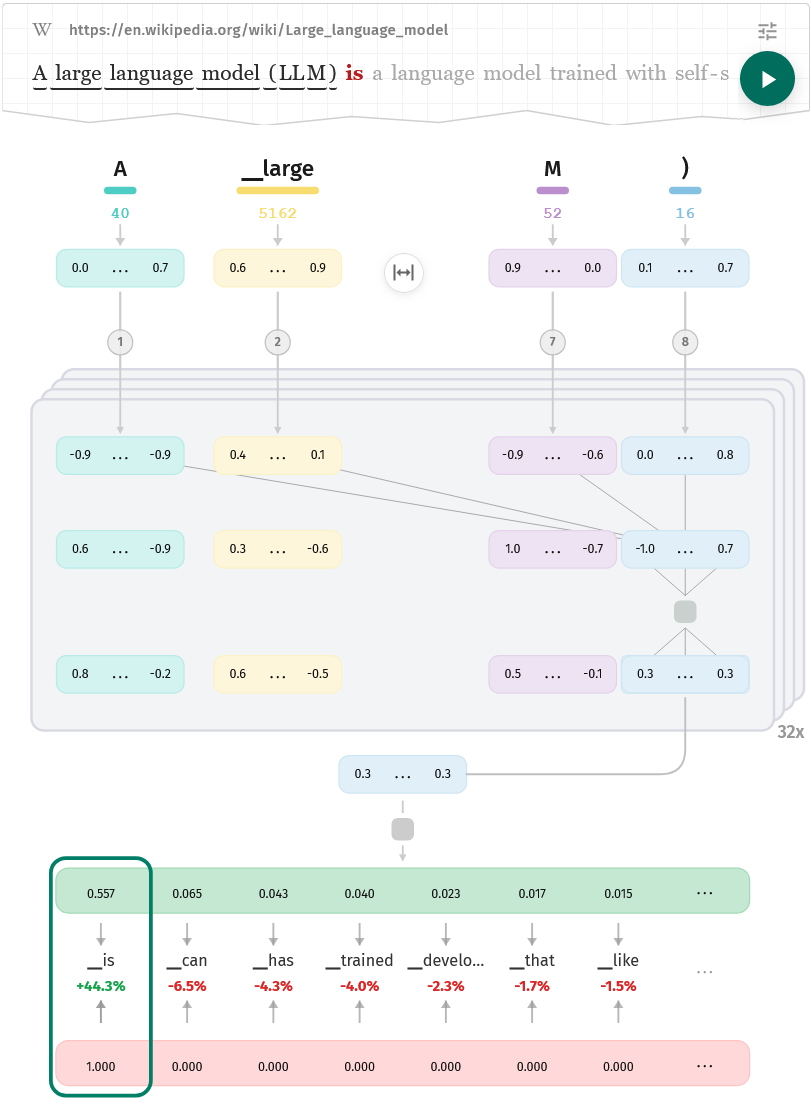}
  \caption{Detailed visualization explaining how the Transformer model is pre-trained on data. Individual components show how the tokens get processed throughout the model.}
  \label{fig:training}
\end{figure}

\section{AnimatedLLM}

\textsc{AnimatedLLM} is a static React\footnote{\url{https://react.dev}} web application that can be hosted on any standard web server.

\begin{figure}[t]
  \centering
  \includegraphics[width=\columnwidth]{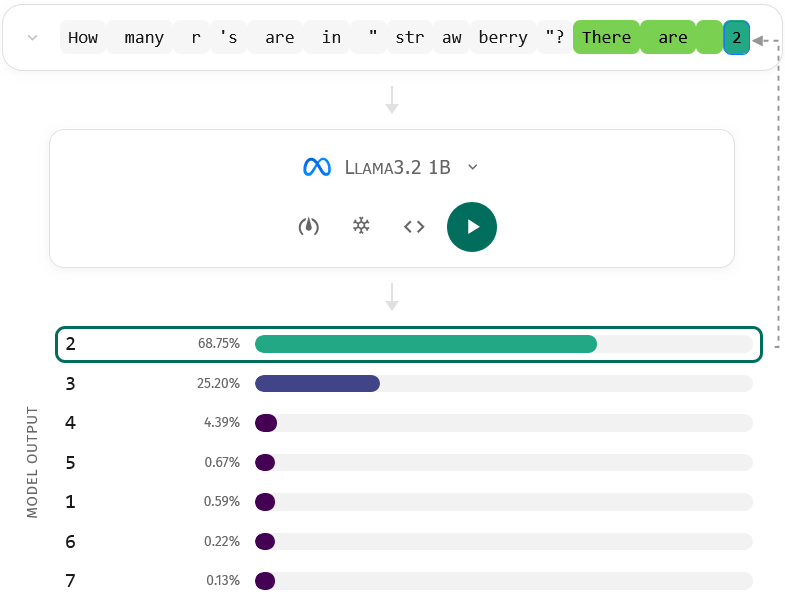}
  \caption{Simplified text generation visualization showing the autoregressive decoding process with probability distributions over candidate tokens.}
  \label{fig:decoding}
\end{figure}

\paragraph{Data} Our goal is to show the inner workings of LLMs using realistic, pedagogically interesting examples. To that end, we pre-computed model traces, i.e., full model outputs for preselected prompts, including tokenization and output probability distributions, on a series of manually-selected prompts. The traces are serialized into JSON files so that the web application can run entirely in the client's browser. We also provide Python scripts for computing traces with custom models and prompts.

\paragraph{Models}  We precomputed traces for multiple models from Huggingface Transformers.\footnote{\url{https://huggingface.com/models}}  We selected the models to cover a variety of providers, model sizes, and pre-training objectives, which results in varied outputs: Olmo 3~\cite{olmo3_report}, Aya Expanse~\cite{dang2024aya}, Llama 3.2~\cite{dubey2024llama}, Qwen 3~\cite{yang2025qwen3}, and GPT-2~\cite{radford2019language}.


\paragraph{Views} \textsc{AnimatedLLM} currently supports four views, each designed for a specific learning objective:
\begin{enumerate}
    \item \textbf{Text generation -- simple view} (see \Cref{fig:decoding}) helps to explain how the next token is selected by demonstrating the principles of autoregressive decoding.
    \item \textbf{Text generation -- detailed view} (see \Cref{fig:full-ui}) shows how the Transformer gradually transforms the input embeddings into the output probabilistic distribution.
    \item \textbf{Pre-training -- simple view}  (see \Cref{fig:training-simple}) helps to explain the training objective: how the model's prediction is compared against the target token.
    \item \textbf{Pre-training -- detailed view}  (see \Cref{fig:training}) relates the Transformer architecture to the training process and helping to understand the principles of backpropagation.
\end{enumerate}
Each animation is accompanied by explanatory labels and tooltips. The tool is designed easily extensible by re-using React components and pre-computed traces. In the future, we plan to implement more views (showing, e.g., post-training approaches or BPE tokenizer training).

\paragraph{User Interface} The web interface follows basic design principles, with the aim of being clean but user-friendly \cite{krug2014dont,williams2015non}. In each view, the user selects an input and starts the animation showing a specific LLM aspect. The user can play, pause, and step through the animation. Keyboard shortcuts enable advanced control (e.g., arrows to step, \texttt{N} to skip to processing the next token, \texttt{G} to jump to the end of the process).  The application can be viewed both on desktop and mobile devices.

\paragraph{Languages} \textsc{AnimatedLLM} is built with multilinguality in mind. The interface is localized into multiple languages, and we provide inputs (along with the associated traces) localized for the language of the interface. Internationalization makes the app more accessible for non-English speakers and opens the possibility to show differences between processing individual languages, for example with respect to tokenization \cite{petrov2023language}. Our application currently supports English, Czech, French, Ukrainian, and Chinese. We are open to community contributions towards translations and prompts for new languages.

\section{Applications}

\textsc{AnimatedLLM} primarily targets non-expert audience that seeks to learn the technical working of LLMs. For example, teachers of high-school computer science or university-level NLP courses can use the tool to provide students with hands-on experience in understanding model behavior. The tool is also available online, so anyone interested in understanding how language models work can explore the visualizations at their own discretion.

We tested preliminary versions of \textsc{AnimatedLLM} at outreach events aimed at high school students with diverse technical backgrounds, including the faculty Open Day\footnote{\url{https://www.mff.cuni.cz/en/admissions/open-days/2025}} and Czech AI Days.\footnote{\url{https://www.dny.ai/event/ai-for-talents-i}} The feedback on the app was generally positive and helped us to ground the discussions with the students. We are currently planning to introduce the application to students in pedagogy programs for high-school computer science education, as well as use it in our own teaching of computer science courses on LLMs and chatbots.

\section{Related Work}
Transformer Explainer \cite{cho2024transformer} is an in-depth interactive visualization of the next-token prediction in the GPT-2 model \cite{radford2019language}. Similarly, LLM Visualization \cite{bycroftllm} provides an interactive walkthrough through individual matrix multiplications inside the Transformer decoder architecture. Build An LLM \cite{buildanllm} is an interactive tool that provides a step-by-step walkthrough for training own LLM locally. Many other visualization tools exist in the LLM interpretability field \cite{brasoveanu2024visualizing}. All of these tools target researchers or more knowledgeable audiences. For lay users, understanding LLMs is facilitated by online blogposts \citep{alammar2018illustrated,alammar2019illustrated} and videos \cite{3blue1brown}.

\section*{Acknowledgments}
This work was funded by the European Union (ERC, NG-NLG, 101039303) and supported by Charles University Research Centre program No.~24/SSH/009. It used resources of the LINDAT/CLARIAH-CZ Research Infrastructure (Czech Ministry of Education, Youth, and Sports project No.~LM2023062). 

\bibliography{custom}

\appendix

\section{App Interface}
\begin{figure*}[h]
  \centering
  {\setlength\fboxsep{0pt}\setlength\fboxrule{0.5pt}\fcolorbox{gray!40}{white}{\includegraphics[width=\linewidth]{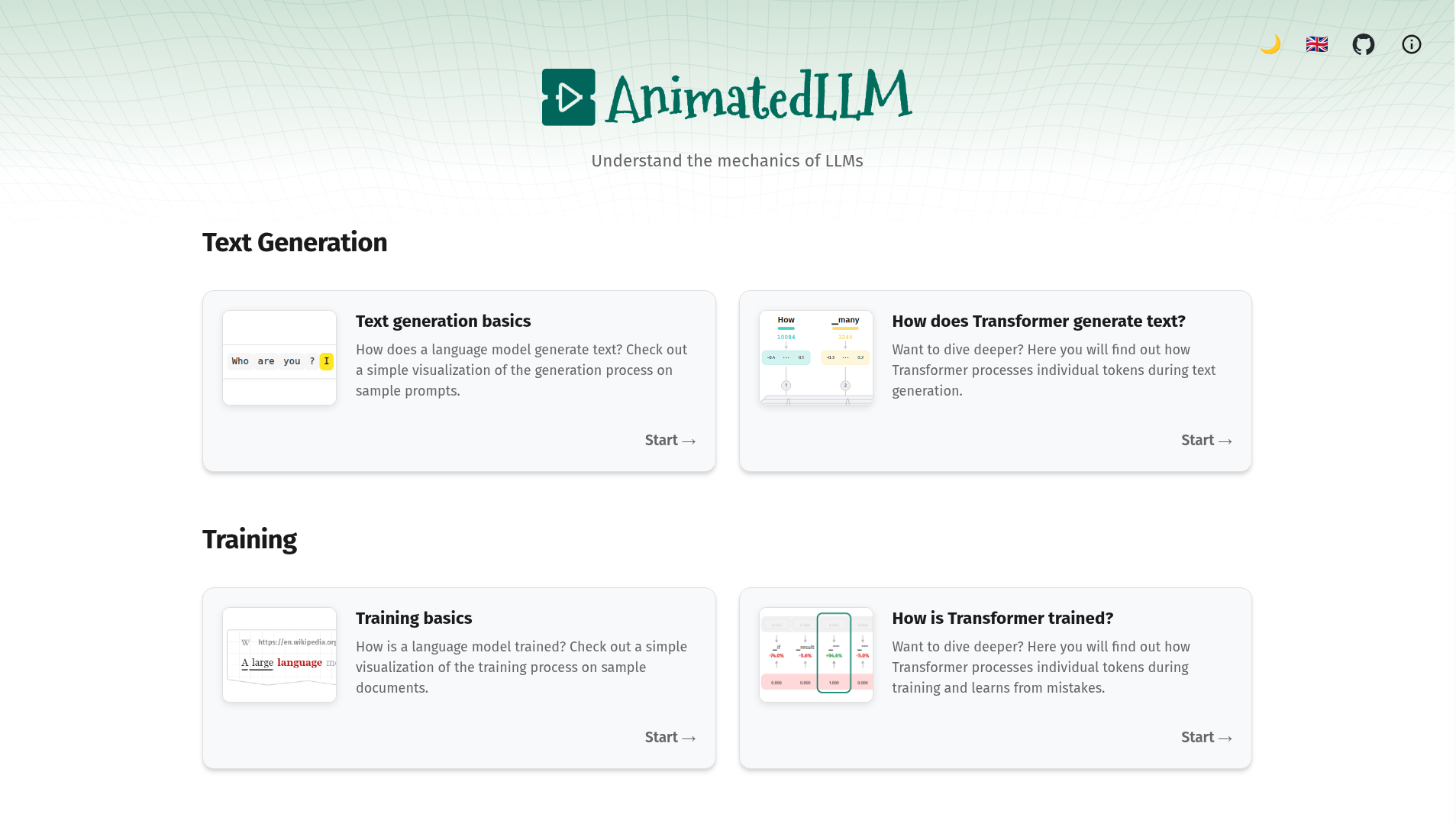}}}
  \caption{Front page of the website providing access to the individual views.}
  \label{fig:front-page}
\end{figure*}

\begin{figure*}[h]
  \centering
  {\setlength\fboxsep{0pt}\setlength\fboxrule{0.5pt}\fcolorbox{gray!40}{white}{\includegraphics[width=\linewidth]{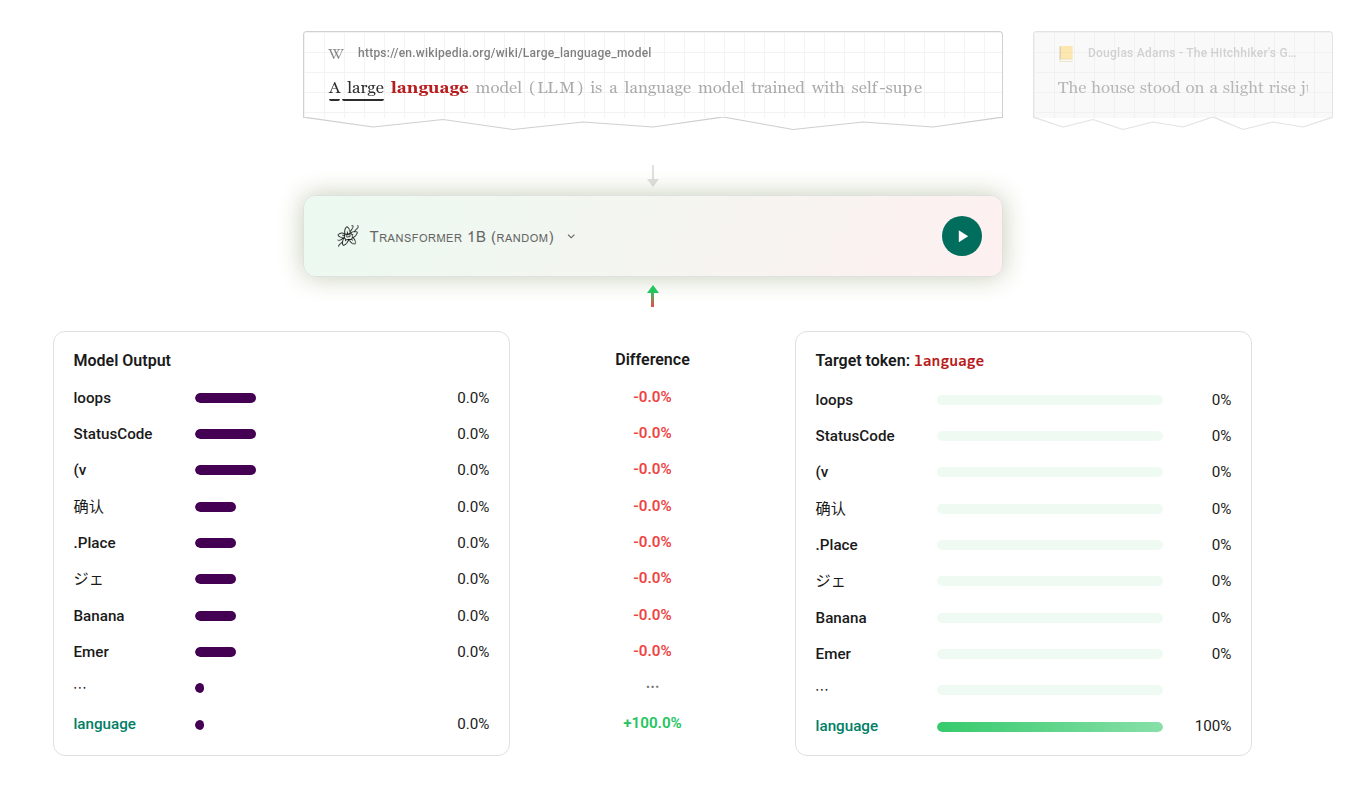}}}
  \caption{Simplified view of the pre-training process, focusing on comparing the predicted and target distribution.}
  \label{fig:training-simple}
\end{figure*}

\begin{figure*}[h]
  \centering
  {\setlength\fboxsep{0pt}\setlength\fboxrule{0.5pt}\fcolorbox{gray!40}{white}{\includegraphics[width=\linewidth]{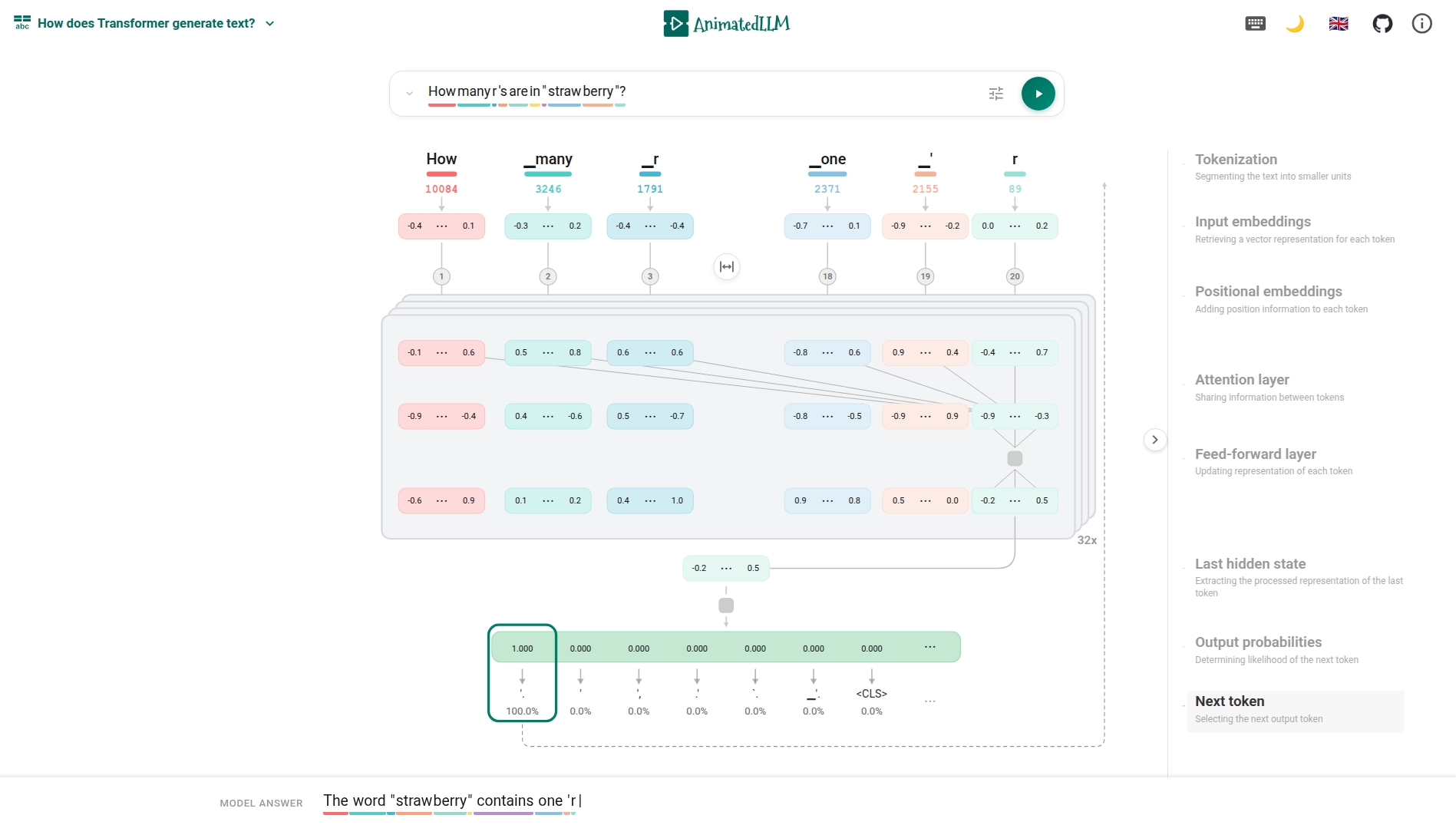}}}
  \caption{The user interface during the detailed visualization of the decoding process. The labels on the side help to understand what is happening in the current animation step.}
  \label{fig:full-ui}
\end{figure*}

\begin{figure*}[h]
  \centering
  {\setlength\fboxsep{0pt}\setlength\fboxrule{0.5pt}\fcolorbox{gray!40}{white}{\includegraphics[width=\linewidth]{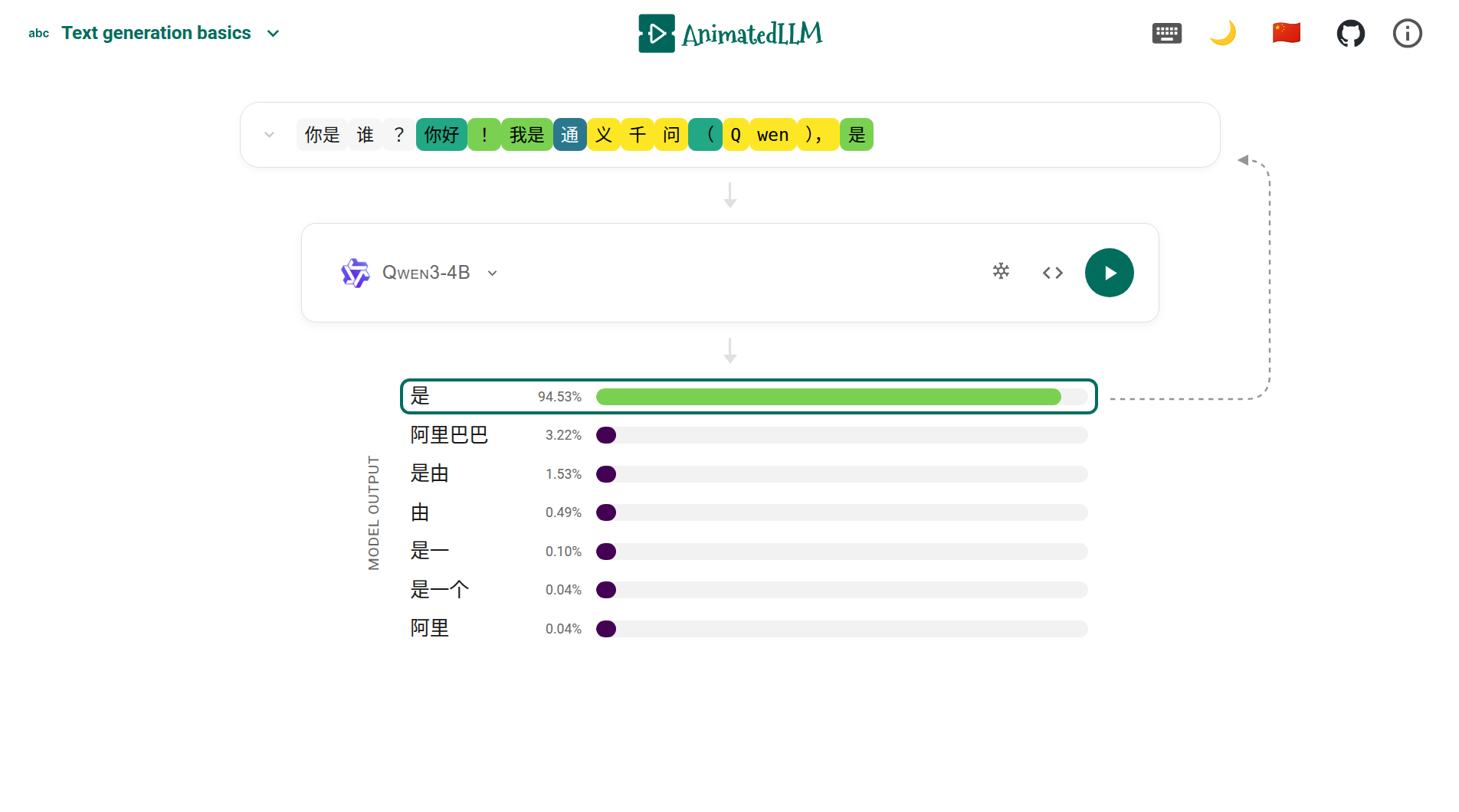}}}
  \caption{\textsc{AnimatedLLM} is localized to several languages, which makes the app more accessible and allows to demonstrate differences between the languages.}
  \label{fig:languages}
\end{figure*}

Additional screenshots of the application interface are shown in \Cref{fig:front-page,fig:training-simple,fig:full-ui,fig:languages}:
\begin{itemize}
  \item \Cref{fig:front-page}: The main page providing access to different visualization views.
  \item \Cref{fig:training-simple}: Simplified pre-training view.
  \item \Cref{fig:full-ui}: Detailed visualization of the decoding process.
  \item \Cref{fig:languages}: Multilingual interface support.
\end{itemize}

\end{document}